\documentclass[conference]{IEEEtran}

\usepackage{amsmath,amsfonts,amssymb,enumerate,amsthm}
\usepackage{bbm}
\usepackage{latexsym}
\usepackage{graphicx}
\usepackage{epstopdf}
\usepackage[margin=0.5cm]{subcaption}
\usepackage{hyperref}
\usepackage{tikz, pgfplots}
\usepackage{epstopdf}
\usepackage{algpseudocode,algorithm}
\usepackage{multirow}
\usepackage{flushend}
\usepackage{rotating}

\newcommand{\argmin}{\operatornamewithlimits{argmin}}

\ifCLASSINFOpdf
\else

\fi
\begin{document}

\title{Classification via Tensor Decompositions \\ of Echo State Networks}

\author{\IEEEauthorblockN{Ashley Prater}
\IEEEauthorblockA{Air Force Research Laboratory\\ Information Directorate \\ Rome, NY 13440 \\ ashley.prater.3@us.af.mil}}

\maketitle

\begin{abstract}
This work introduces a tensor-based method to perform supervised classification on spatiotemporal data processed in an echo state network.  
Typically when performing supervised classification tasks on data processed in an echo state network, the entire
collection of hidden layer node states from the training dataset is shaped into a matrix, allowing one to use standard linear algebra techniques to train the output layer.  
However, the collection of hidden layer states is multidimensional in nature, and representing it as a matrix may lead to undesirable numerical conditions or loss of spatial and temporal correlations in the data. 

This work proposes a tensor-based supervised classification method on echo state network data that preserves and exploits the multidimensional nature of the hidden layer states.  The method, which is based on orthogonal Tucker decompositions of tensors, is compared with the standard linear output weight approach in several numerical experiments on both synthetic and natural data.   The results show that the tensor-based approach tends to outperform the standard approach in terms of classification accuracy.

\end{abstract}

\IEEEpeerreviewmaketitle
\thispagestyle{empty}

\section{Introduction}
Echo State Networks (ESNs), first introduced in~\cite{jaeger} and~\cite{maasrealtime} under the name \emph{Liquid State Machines}, have been shown to be effective at performing classification and temporal prediction tasks on spatiotemporal data, including such diverse tasks as speech recognition~\cite{maasrealtime, skow, verstraeten2, jaegerleaky, prater_NN}, chaotic time-series prediction~\cite{jaeger_lstm}, and forming the objective function in reinfocement learning methods~\cite{bush, szita}.
ESNs are a special type of recurrent neural network (RNN). 
Typically the hidden layer in an RNN is trained using a computational expensive back propagation method~\cite{werbos}.  In ESNs however, the weights of the hidden layer are randomly assigned with sparse, random connections among the nodes. The hidden layer of an ESN is often called a \emph{reservoir}, with the collection of node values called the \emph{reservoir states}.  The advantages ESNs offer over traditional RNNs include a much faster training time, and a configuration that does not require retraining to use in new applications on different datasets.

In ESNs, only the output layer is trained for a particular task, and the training generally produces linear output weights using a regularized least squares approach~\cite{goudarzi,jaeger, luko2, luko}.  Recently, the output layer of an ESN was replaced with a classification scheme based on the principal components of the reservoir states~\cite{prater_IJCNN, prater_NN}. This approach showed promise in improving classification accuracy and in being more robust to noisy perturbations in the input data than the traditional linear trained output weights approach.  However, the collection of reservoir states generated by an ESN is multidimensional in nature.  Both trained linear output weights and the principal components approaches require one to superficially flatten the reservoir data into matrices, potentially eliminating or weakening spatial and temporal correlations present in the data in the process.  Additionally, flattening the reservoir data may result in a very overdetermined system, especially for the trained linear output weights approach, which may yield overtrained or overly sensitive results.  For example, see~\cite{prater_IJCNN}.

To this end, this work proposes a tensor-based supervised classification method for use with reservoir states of ESNs.  Tensors, or multidimensional arrays, are natural structures for representing collections of reservoir data.  Rather than using the raw reservoir data, the tensors will be approximated using a Tucker decomposition~\cite{tucker}, with each mode having a factor matrix of smaller rank 
along with an associated core tensor that extracts the higher order features of the data.  This decomposition enables one to reduce the complexity of the training tensor data by keeping only the most significant contributions while preserving the multidimensional correlations among the features.

The following notation appears in this work.  Tensors will be written in a script capital letters, e.g.\ $\mathcal{A}$, matrices will appear as capital latin letters, e.g.\  $A$, and vectors and scalars will be lower case latin or greek letters, e.g.\ $a$ or $\alpha$.  
Elements of an array will be given in `matlab' notation.  Thus the $(i,j)^\text{th}$ element of matrix $A$, that is the element in the $i^\text{th}$ row and $j^\text{th}$ column, will be denoted by
$A(i,j)$.  The vector determined by extracting the $j^\text{th}$ column from $A$ is denoted by $A(:,j)$.  Similar notation holds for lower or higher-order arrays.
The usual matrix-matrix multiplication will be represented by writing the matrices adjacent to one another.  This is in contrast to modal tensor-matrix multiplication, which will be defined in Section~\ref{sec:background}.  Superscripts will denote indices in a set, not power-wise multiplication, except for when the base is a set as in $\mathbb{R}^N$.  The Hermitian transpose of a matrix $A$ is given by $A'$. Finally,
$e_k$ will denote a vector, of length clear from context, with a `1' in the k-th position and zeros elsewhere. 

The remainder of this paper is organized as follows.  Section~\ref{sec:background} discusses background information for ESNs and relevant tensor decompositions.  Section~\ref{sec:method} describes the proposed classification method using tensor decompositions on the reservoir states.  The results of several numerical experiments are presented in Section~\ref{sec:experiments}.  Finally, Section~\ref{sec:conclusion} contains conclusions and a dicussion of future work.

\section{Background}\label{sec:background}
In this section, background information is given on tensor decompositions and echo state networks.

\subsection{Tensor Decompositions}

A tensor is a higher-order analogue of a vector; A vector is a first-order tensor, a matrix is a second-order tensor, and so on.  Tensors are natural structures to represent and investigate multidimensional data.  For example, video may be considered a third-order tensor, with the first two modes describing the $x$ and $y$ pixel coordinates of a single frame, and the third mode representing time variations.  Tensors have been used to represent and explore relationships among data in diverse research areas and applications, including video processing~\cite{li}, multiarray signal processing~\cite{lim}, independent component analysis~\cite{goyal}, and others.

One challenge to employing tensor methods is the volume of the data, which suffers from the so-called `curse of dimensionality'~\cite{bellman_curse}.  That is, the amount of data increases exponentially with each additional mode, and naive tensor methods quickly become intractable.  To help alleviate this challenge various approaches have been proposed, including several types of low-rank and sparse representations~\cite{shah}. In this work, we will employ the orthogonal Tucker-2 decomposition.

Before discussing the decomposition, some definitions need to be introduced.  Let ${\mathcal{A} \in \mathbb{R}^{I_1\times I_2\times \cdots \times I_N}}$ be an $N^\text{th}$-order tensor, with $n^\text{th}$ mode having dimension $I_n$, and let ${B\in\mathbb{R}^{J_n\times I_n}}$ be a matrix.  The \emph{$n^\text{th}$ mode product of $\mathcal{A}$ by $B$} is the $n^\text{th}$-order tensor ${\mathcal{A}\times_n B \in \mathbb{R}^{I_1\times I_2\times \cdots\times I_{n-1}\times J_n \times I_{n+1} \times \cdots \times I_N}}$ whose entries are given by
\begin{align*}
	(\mathcal{A}\times_n B)&(i_1,i_2,\ldots,i_{n-1},j_n,i_{n+1},\ldots,i_n) = \\
	&= \sum_{i_n=1}^{I_n} \mathcal{A}(i_1,i_2,\ldots,i_N) B(j_n,i_n)
\end{align*}
\cite{de_lathauwer, kolda, phan}.

Modal products of a tensor by matrices are commutative, provided the modes are distinct.  That is, given the tensor $\mathcal{A}\in\mathbb{R}^{I_1\times I_2\times \ldots \times I_N}$ and the matrices $B\in\mathbb{R}^{J_n\times I_n}$ and $C\in\mathbb{R}^{J_m\times I_m}$, then
\begin{equation*}
	\left(\mathcal{A}\times_n B\right) \times_m C = \left(\mathcal{A}\times_m C\right)\times_n B = \mathcal{A}\times_n B \times_m C,
\end{equation*}
provided $m\neq n$.

A tensor may be represented as a matrix through the process of unfolding~\cite{kolda, kroonenberg}, similar to the way a matrix may be represented as a vector by vectorization.  
The \emph{matrix unfolding of $\mathcal{A}$ in the third mode} is the matrix $\mathcal{A}_{(3)} \in \mathbb{R}^{I_1\times (I_2 I_3)}$ with elements 
\begin{equation*}
	\mathcal{A}_{(3)}(i_1,(i_3-1)I_2 + i_2) = \mathcal{A}(i_1,i_2,i_3).
\end{equation*}
One may think of $\mathcal{A}_{(3)}$ as the concatenated matrix
\[ \mathcal{A}_{(3)} = \begin{bmatrix} \mathcal{A}(:,:,1) | \mathcal{A}(:,:,2) | \cdots | \mathcal{A}(:,:,I_3)\end{bmatrix}. \]
Unfoldings of higher-order tensors or in different modes follow an analogous procedure.

An inner product may be defined on tensors as follows.  Let 
\begin{equation*}
	\langle \cdot, \cdot \rangle : \mathbb{R}^{I_1\times \cdots \times I_N} \times \mathbb{R}^{I_1\times \cdots \times I_N}  \rightarrow \mathbb{R}
\end{equation*}
be defined by 
\begin{equation*}
	\langle \mathcal{A} , \mathcal{B}\rangle = \sum_{i_1,\ldots,i_N} \mathcal{A}(i_1,\ldots,i_N) \mathcal{B}(i_1,\ldots,i_N).
\end{equation*}
This inner product induces the tensor Frobenius norm, 
\begin{equation*}
	\left\| \mathcal{A} \right\| = \sqrt{ \langle \mathcal{A},\mathcal{A} \rangle }.
\end{equation*}

Now equipped with the requisite definitions, we are ready to build the orthogonal Tucker-2 tensor decomposition.  

The Tucker decomposition expresses a tensor as the modal product of a core tensor by several matrices~\cite{cichocki, kolda, tucker}.  For a third-order tensor $\mathcal{A}^{I_1\times I_2 \times I_3}$, the Tucker decomposition is
\begin{equation}\label{eq:tucker}
	\mathcal{A} = \mathcal{B} \times_1 A \times_2 B \times_3 C,
\end{equation}
where $A\in\mathbb{R}^{J_1\times I_1}, B\in\mathbb{R}^{J_2\times I_2}$ and $C\in\mathbb{R}^{J_3\times I_3}$ are the factor matrices, typically with $J_n \ll I_n$, and $\mathcal{B}\in\mathbb{R}^{J_1\times J_2\times J_3}$ is the \emph{core tensor}.  If the matrices $A, B$ and $C$ each have orthogonal columns, then~\eqref{eq:tucker} is called an \emph{orthogonal tucker} decomposition, and is expressed as 
\begin{equation}\label{eq:orthog tucker}
	\mathcal{A} = \mathcal{B}\times_1 U \times_2 V \times_3 W.
\end{equation}

\begin{figure*}
	\centering
	\begin{tikzpicture}
	\draw (0,0) -- (2,0) -- (2,3) -- (0,3) -- (0,0);
	\draw (0,3) -- (1.5,4.5) -- (3.5,4.5) -- (2,3);
	\draw (3.5,4.5) -- (3.5,1.5) -- (2,0);
	%
	\node at (4,2) {$=$};
	%
	\draw (5,0) -- (6,0) -- (6,3) -- (5,3) -- (5,0);
	%
	\draw (6.5,3) -- (6.5,2) -- (8,2) -- (8,3) -- (6.5,3);
	\draw (6.5,3) -- (8,4.5) -- (9.5,4.5) -- (8,3);
	\draw (9.5,4.5) -- (9.5,3.5) -- (8,2);
	%
	\draw (10,3) -- (10,1.5) -- (12,1.5) -- (12,3) -- (10,3);
	%
	\node at (1,1.5) {$\mathcal{A}$};
	\node at (5.5,1.5) {$U^\top$};
	\node at (7.25,2.5) {$\mathcal{F}$};
	\node at (11,2.25) {$V$};
	\end{tikzpicture}
\caption{An illustration of the two-way orthogonal Tucker decomposition~\eqref{eq:tucker2}.}
\label{fig:tucker2}
\end{figure*}
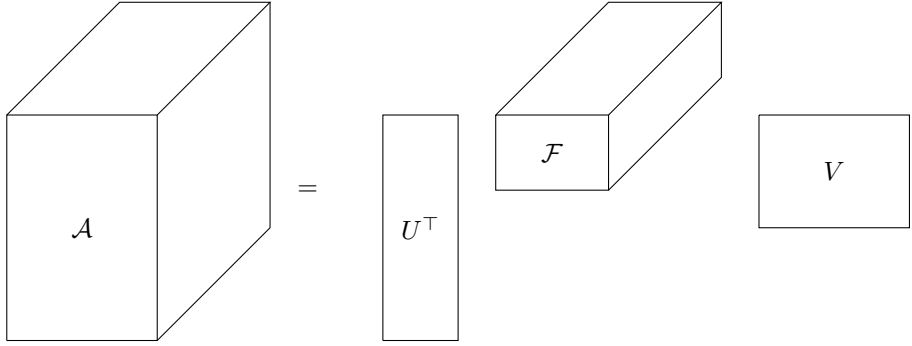

A variant of the orthogonal Tucker decomposition, called the Tucker-2 decomposition, is of the form
\begin{equation}\label{eq:tucker2}
	\mathcal{A} = \mathcal{F}\times_1 U \times_2 V,
\end{equation}
with core tensor $F\in\mathbb{R}^{J_1\times J_2\times I_3}$.  That is, the original tensor $\mathcal{A}$ of order greater than two is written using only two factor matrices.  An illustration of the decomposition~\eqref{eq:tucker2} is shown in Figure~\ref{fig:tucker2}. 
This type of decomposition was used in~\cite{phan} to perform feature extraction and classification of a collection of 2-dimensional signals, where the third mode of the tensors $\mathcal{A}$ and $\mathcal{F}$ correspond to the individual samples in the training dataset.  Note that in the decomposition~\eqref{eq:tucker2}, the basis matrices $U$ and $V$ are universal across the entire tensor $\mathcal{A}$.  In classification tasks, this method does not generate different basis matrices for distinct classes in the data set.   The decomposition~\eqref{eq:tucker2} and classification methods using it will be discussed further in Section~\ref{sec:method} for use with Echo State Network data.


\subsection{Echo State Networks}

Spatiotemporal data are processed in an echo state network as follows.  Let $A\in\mathbb{R}^{L\times T}$ be an input of spatial dimension $L$ and temporal length $T$.  The values of the hidden layer nodes, also referred to as the \emph{reservoir states} of the input $A$, denoted by  $X \in \mathbb{R}^{N\times T}$, are determined by the recursion
\begin{align}\label{eq:ESN}
	X&(:,t+1) =  \\
	&=(1-\alpha)X(:,t) + \alpha f\left( W_\text{in} A(:,t) + W_\text{res}X(:,t) + \beta \right). \nonumber 
\end{align}

In~\eqref{eq:ESN}, $f$ is a nonlinear activation function, ${W_\text{in} \in \mathbb{R}^{N\times L}}$ are the fixed input weights, ${W_\text{res} \in \mathbb{R}^{N\times N}}$ are the fixed, randomly assigned reservoir weights, $\beta$ is a bias value, and $\alpha \in [0,1]$ is the leaking rate.  This work does not use output feedback in the ESN recursion as is sometimes used in literature~\cite{jaeger}, since it is incompatible with using the proposed tensor approach. 

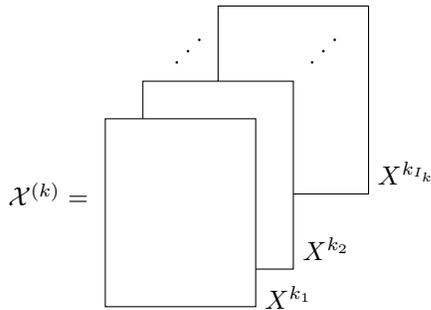
\begin{figure}[bhpt]
\centering
\begin{tikzpicture}
	\draw (2,0.5) rectangle (4,3);
	\draw (2.5,3) -- (2.5,3.5) -- (4.5,3.5) -- (4.5,1) -- (4,1);
	\draw (3.5,3.5) -- (3.5,4.5) -- (5.5,4.5) -- (5.5,2) -- (4.5,2);
	\draw (3.10,3.90) circle (0.01);
	\draw (3.25,4.05) circle (0.01);
	\draw (2.95,3.75) circle (0.01);

	\draw (4.90,3.90) circle (0.01);
	\draw (5.05,4.05) circle (0.01);
	\draw (4.75,3.75) circle (0.01);

	\node at (4,0.35) [anchor = south west] {$X^{k_1}$};
	\node at (4.5,1) [anchor = south west] {$X^{k_2}$};
	\node at (5.5,2) [anchor = south west] {$X^{k_{I_k}}$};

	\node at (1.9,2) [anchor=east] {$\mathcal{X}^{(k)} = $};
\end{tikzpicture}
\caption{A representation of the tensor $\mathcal{X}^{(k)}$, whose frontal slices are collections of reservoir states from the $k^\text{th}$ collection of training inputs.}
\label{fig:ESN tensor}
\end{figure}

For supervised classification tasks, one will have a collection of training samples.  Suppose the training data are partitioned into $K$ classes, and the $k^\text{th}$ class contains $I_k$ examples. Say 
\begin{equation*}
	\mathrm{Tr} := \left\{ A^1, A^2, \ldots \right\}
\end{equation*}
 is the entire collection of training inputs and 
 \begin{equation*}\label{eq:training samples}
	\mathrm{Tr}_k := \left\{ A^{k_1}, A^{k_2}, \ldots, A^{k_{I_k}} \right\}
\end{equation*}
is the collection of training inputs from the $k^\text{th}$ class.   Suppose each of the inputs are processed in the ESN~\eqref{eq:ESN} with the same weights and parameters.  Denote the reservoir states from the input $A^{k_j}$ as $X^{k_j}$. 
All of the reservoir states from $\mathrm{Tr}_k$ may be concatenated along the third mode into the tensor ${\mathcal{X}^{k} \in \mathbb{R}^{N\times T\times |\mathrm{Tr}_k|}}$, where
\begin{equation*}
	\mathcal{X}^k(:,:,j) = X^{k_j}.
\end{equation*}
A representation of the tensor $\mathcal{X}^{k}$ is shown in Figure~\ref{fig:ESN tensor}.  Similarly, all of the reservoir states may be concatenated along the third mode into the tensor $\mathcal{X} \in \mathbb{R}^{N\times T \times |\mathrm{Tr}|}$.

Traditionally, a collection of linear output weights is trained on the unfolded collection of training reservoir tensors.  Let $X_{(3)} \in \mathbb{R}^{N\times T|\mathrm{Tr}|}$ be the unfolding of $\mathcal{X}$ along the third mode.  That is, $X_{(3)}$ may be written as the contatenation
\begin{equation*}
	X_{(3)} = \begin{bmatrix} \mathcal{X}(:,:,1) | \mathcal{X}(:,:,2) | \cdots | \mathcal{X}(:,:,|\mathrm{Tr}|)\end{bmatrix}.
\end{equation*}
Let $y \in \mathbb{R}^{K \times T|\mathrm{Tr}|}$ be a matrix whose columns indicate class membership.  Say the columns of $y$ satisfy ${y(:,(j-1)T + t) = e_k}$.

In the standard approach~\cite{jaeger} to performing classification using ESNs, first ones finds a collection of output weights ${W_\text{out} \in \mathbb{R}^{K\times N}}$ so that 
\begin{equation}\label{eq:Wout}
	W_\text{out} X_{(3)} \approx y.
\end{equation}
Commonly a solution to~\eqref{eq:Wout} is found using a regularization method, such as
\begin{equation}\label{eq:RLS}
	W_\text{out} = \argmin_{W \in \mathbb{R}^{K\times N}} \left\{\left\|WX_{(3)} - y\right\|_2^2 + \lambda \|W\|_2^2\right\}.
\end{equation} 
If a new input $A$ belongs to the $k^\text{th}$ class at time $t$, and $W_\text{out}$ describes the data well, then $W_\text{out}A(:,t) \approx e_k$.  This observation drives the classification scheme.  Say the input $A$ from the test set with reservoir states $X$ is predicted to belong to class $k$ at time $t$ if the the maximal element of the vector $W_\text{out}X(:,t)$ is in the $k^\text{th}$ entry.  Similarly, $A$ is predicted to belong to the $k^\text{th}$ class overall if the vector 
\begin{equation}\label{eq:Wout class}
	\sum_t W_\text{out} X(:,t)
\end{equation}
 is maximized in the $k^\text{th}$ entry.
Note that a single output weight matrix~\eqref{eq:RLS} is found and applied to all test samples at all time steps.  

Training output weights as in~\eqref{eq:RLS} is fast, however this approach does have weakensses.  The system~\eqref{eq:Wout} is typically overdetermined.  Although the regularization~\eqref{eq:RLS} tries to overcome this, the matrix $X_{(3)}$ may have several orders of magnitude more columns than rows in practice and may poorly represent the data, even with regularization.  This may be controlled by the model by selecting only a subset $\Omega \subset \{1,2,\ldots,T\}$ of times at which to sample the reservoir.  One common method is to use only a single point $\Omega = \{T\}$~\cite{jaeger}, however the accuracy results may suffer greatly~\cite{prater_IJCNN}.  Even though the reservoir states hold some `memory' of previous states, using only a subset of the data generally results in information loss.  
Moreover, the unfolding procedure loses some temporal correlations of reservoir node behavior.  Finally, using a single linear output weight may simply be insufficient to separate the classes in the dataset well, as shown in~\cite{prater_NN}.

\section{Tensor Decompositions of Reservoir States}\label{sec:method}
To alleviate the deficiencies encountered using method~\eqref{eq:Wout class}, this paper proposes using a classification method based on the decomposition of the tensor of training reservoirs.   To this end, let approximations of the orthogonal Tucker decompositions of the tensors $\mathcal{X}$ and $\mathcal{X}^{k}$ be 
\begin{equation}\label{eq:tuckerX}
	\mathcal{X} \approx \mathcal{F} \times_1 U \times_2 V
\end{equation} 
and 
\begin{equation}\label{eq:tuckerXk}
	\mathcal{X}^k \approx \mathcal{F}^k \times_1 U^k \times_2 V^k
\end{equation}
where the factors ${U, U^k \in \mathbb{R}^{N\times J_1}}, {V, V^k \in \mathbb{R}^{T\times J_2}}$ have orthogonal columns with $J_1\ll N$ and $J_2\ll T$, and ${\mathcal{F}\in\mathbb{R}^{J_1\times J_2\times |\mathrm{Tr}|}}$ and ${\mathcal{F}^k\in\mathbb{R}^{J_1\times J_2\times \mathrm{Tr}_k}}$ are core tensors.    The core tensors may be intrepreted as the entry $\mathcal{F}(i_1,i_2,j)$ describing the strength of the feature in the reservoir states of the $j^\text{th}$ input captured by the interaction of the bases $U(:,i_1)$ and $V(:,i_2)$~\cite{phan}.

To approximate the Tucker-2 decomposition of the form~\eqref{eq:tuckerX}, we use the Higher-Order Orthogonal Iteration (HOOI) algorithm, first introduced in~\cite{de_lathauwer} and explored in~\cite{phan}.  For completeness, we include the pseudocode as Algorithm~\ref{alg:HOOI}.  A similar procedure can be used to obtain the decompositions~\eqref{eq:tuckerXk}.

\begin{algorithm}
\caption{HOOI}\label{alg:HOOI}
\begin{algorithmic}[1]
\State {\bf Inputs:} Tensor of reservoir states from the training set $\mathcal{X}$; Factor ranks $J_1, J_2$; $tol>0$.
\State {\bf Outputs:} Factor matrices $U, V$ and core tensor $\mathcal{F}$.
\State {\bf Initialization:} Let $n = 0$.  Randomly choose basis factor matrices $U^{0}$ and $V^0$.
\While {$\max\{\| S^n_1 - S_1^{n+1}\|, \| S^n_2 - S_2^{n+1}\|\} \geq tol$}
\State \emph{Update mode-1 factor:}
\State $\mathcal{B} = \mathcal{X}\times_2 {(V^n)}'$
\State $[u,s,v] = \mathrm{svd}(\mathcal{B}_{(1)})$
\State $U^{n+1} = u(:,1:J_1), \; S_1^{n+1} = \mathrm{diag}(s)$
\State
\State \emph{Update mode-2 factor:}
\State $\mathcal{B} = \mathcal{X}\times_2 {(U^{n+1})}'$
\State $[u,s,v] = \mathrm{svd}(\mathcal{B}_{(2)})$
\State $V^{n+1} = u(:,1:J_2), \; S_2^{n+1} = \mathrm{diag}(s)$
\State
\State $n \leftarrow n+1$
\EndWhile
\State $U = U^\text{end},\; V = V^\text{end}$
\State $\mathcal{F} = \mathcal{X} \times_1 U' \times_2 V'$
\end{algorithmic}
\end{algorithm}


In~\cite{de_lathauwer}, the factor matrices $U^{(0)}, V^{(0)}$ are initialized as the dominant left singular subspace of the unfolded tensors $\mathcal{X}_{(1)}$ and $\mathcal{X}_{(2)}$.  In this work, we randomly initialize them for two reasons.  First, just finding the dominant subspaces of the unfolded matrices is a very computationally intensive task if $\mathcal{X}$ is large.  In practice, this step may be intractable, even if steps 6 and 11 in Algorithm~\ref{alg:HOOI} are computable.  Second, it was noted in experiments that randomly initializing the factors results in only a few additional iterations. 
Algorithm~\ref{alg:HOOI} uses a stopping criterion based on the convergence of the singular values found in Steps 6 and 11, this is to avoid problems in difference in signs when using a criterion based on the factor matrices. 


One may perform classification using the decompositions~\eqref{eq:tuckerX} and~\eqref{eq:tuckerXk} obtained from collections of reservoir states of a training set.  To do so, suppose $A$ is a new input signal with reservoir states $X$.  Although $X$ is a matrix rather than a three dimensional tensor, it can still be expressed in terms of the factor matrices from the Tucker decompositions.  Say 
\begin{equation}\label{eq:XG}
	X = G \times_1 U \times_2 V
\end{equation}
and
\begin{equation}\label{eq:XGk}
	X = G^k \times_1 U^k \times_2 V^k
\end{equation}
for each $k$, where $G$ and $G^k$ are the core matrices, and $U, V, U^k, V^k$ are found in the training step. 
Since the factor matrices are orthogonal and $X$ is a matrix, Equations~\eqref{eq:XG} and~\eqref{eq:XGk} may be rewritten as
\begin{equation}\label{eq:G}
	G = X\times_1 U' \times_2 V' = U'XV
\end{equation}
and
\begin{equation}\label{eq:Gk}
	G^k = X \times_1 {U^k}' \times_2 {V^k}' = {U^k}'XV^k.
\end{equation}
Indeed, each frontal slice of $\mathcal{F}$ is of the form~\eqref{eq:G}, with a collection of reservoir states from the training set in place of $X$.  It is reasonable to assume that inputs from the same class have similar reservoir responses, and therefore also produce similar frontal slices in the core tensors.  Therefore, one may predict that an input $A$ belongs to the $k^\text{th}$ class if the slices from $\mathrm{Tr}_k$ describe $G$ well.  That is, say $A$ is in the $k^\text{th}$ class if 
\begin{equation}\label{eq:tensor class 1}
	j = \argmin \left\| G - \mathcal{F}(:,:,j) \right\|
\end{equation}
and $U^j \in \mathrm{Tr}_k$.
Similarly, one could predict $U$ belongs to the $k^\text{th}$ class if 
\begin{equation}\label{eq:tensor class 2}
	k = \argmin_k\left\{ \min_j \left\| G^k - \mathcal{F}(:,:,j) \right\| \right\}.
\end{equation}

\section{Experimental Results}\label{sec:experiments}

In this section, the results of numerical experiments are given, comparing the classification accuracy using ESNs with the linear output weight approach~\eqref{eq:RLS} with the proposed tensor-based classification methods~\eqref{eq:tensor class 1}, and~\eqref{eq:tensor class 2}. 
Three datasets are used.    
The first dataset uses inputs that randomly switch between sine wave and square wave segments.  
The second dataset is a subset of the USPS collection of handwritten digits.  
The final dataset is a collection of cepstrum coefficients from audio recordings of speakers saying the Japanese vowel `ae'.

All experiments are performed in MATLAB 2017a on a PC with 16GB RAM.  Several parameter combinations are considered for each dataset, with experiments repeated a number of times for each combination.  The randomizations in the weights $W_\text{in}$ and $W_\text{res}$  and in generating training and testing datasets are reselected for each experiment, however they are held constant for all classification methods within a single experiment.

\subsection{Sine vs.\ Square Wave}
\begin{figure}[hbpt]
	\centering
	\includegraphics[width=0.45\textwidth]{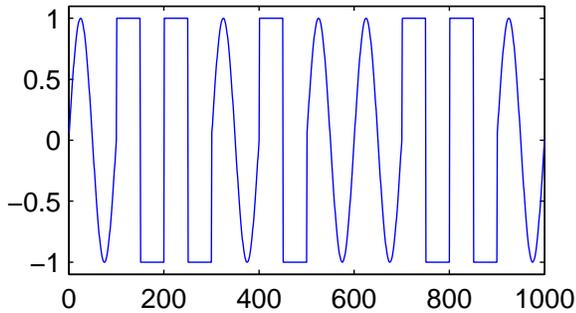}
	\caption{A typical input signal from the `Sine vs.\ Square Wave' example.}
	\label{fig:SineSquareExample}
\end{figure}

In this collection of experiments, the input signals are formed by randomly placing sine and square wave segments of length and period 100,  and paired with an indicator matrix $y\in\mathbb{R}^{2,T}$ where 
\begin{equation*}
	y(:,t) = \begin{cases} \begin{bmatrix} 1 &0 \end{bmatrix}^\top, &\text{ if $u$ is a sine wave at time $t$,} \\ & \\\begin{bmatrix} 0 & 1\end{bmatrix}^\top, &\text{ if $u$ is a square wave at time $t$.} \end{cases}
\end{equation*}
A sample training input is shown in Figure~\ref{fig:SineSquareExample}. A dataset of this type has been studied in previous ESN work, including~\cite{prater_IJCNN} for use with ESN matrix principal component output layer classification methods, and~\cite{paquotopto, zhang} for study using photonic reservoirs.

The training data is formed by generating 20 distinct input patterns, each containing 100 segments of randomly placed sine and square waves.  The test set is generated similarly, but contains 50 distinct patterns.  The ESN parameters used in the simulations are ${N \in \{10, 20, 100\}}, \; {f\in\{\tanh, \sin\}}, {\beta \in \{0,\pi/4\}}$ and ${\alpha = 1}$.  For each ${(N,f,\beta)}$, the experiments are repeated 50 times with new randomizations in $W_\text{res}, \; W_\text{in}$ and the training and testing sets.    For tensor-based classification, we used the method~\eqref{eq:tensor class 2} with ranks ${R_1 = \lfloor N/5\rfloor}, \; {R_2 = 5}, \; {R_3 = 1}$.  The choice $R_3=1$ is justified because the the dataset is rather simple; only one type of square wave and one type of sine wave are used in generating the samples.

For tensor-based classification, the data generated by the ESNs on the training set are partitioned into two tensors $\mathcal{X}^1 \in \mathbb{R}^{N\times 100 \times I_1}$ and $\mathcal{X}^2\in\mathbb{R}^{N\times 100 \times I_2}$, where $\mathcal{X}^1$ contains the reservoir states corresponding to the `sine' inputs, and $\mathcal{X}^2$ contains the reservoir states corresponding to the `square' inputs.  Each tensor is decomposed as in~\eqref{eq:tuckerXk}:
\begin{equation*}
	\mathcal{X}^1 \approx \mathcal{F}^2 \times_1 U^1 \times V^1,
\end{equation*}
and
\begin{equation*}
	\mathcal{X}^2 \approx \mathcal{F}^2 \times U^2 \times V^2.
\end{equation*}
For each test set element, the segments of length 100 are classified according to~\eqref{eq:tensor class 2}. That is, let $A$ be a new input pattern from the test set with reservoir states $X$.  Let ${X_t = X(:,100(t-1)+1:100t)}$ be the collection of states corresponding to the $t^\text{th}$ input segment of $A$.  Say that the $t^\text{th}$ segment is classified as a `sine' wave if 
\begin{align*}
	\min_{j_1} &\left\| X_t \times_1 {U^1}' \times_2 {V^1}'  - \mathcal{F}^1(:,:,j_1) \right\| \leq  \\
	&\leq \min_{j_2} \left\| X_t \times_1{U^2}' \times_2{V^2}' - \mathcal{F}^2(:,:,j_2) \right\|
\end{align*}
and as a `square' wave segment if the inequality sign is flipped.

For trained linear output weights-based classification, we generate a single matrix $W_\text{out}$ via Equation~\eqref{eq:RLS}, and perform classification on the test set both pointwise and block-wise on each input segment as in Equation~\eqref{eq:Wout class}.

The mean and standard deviation of the percent classification accuracy results using these methods over 50 simulations for several parameter choices are displayed in Table~\ref{tab:SineSquareExample}.  The tensor-based classification method, in columns labeled `Tensor', achieved 100\% accuracy in every simulation on both the training and testing datasets.  The pointwise and block-wise output weight classification methods, in columns labeled `Weights (pt)' and `Weights (bk)' respectively, achieved good classification accuracy for some parameter choices, but poor results for others.  The block-wise method is sensitive to the number of nodes in the ESN and the bias choice, and in particular achieved near-perfect test set classification accuracy when $\beta$ was chosen well.  On the other hand, the pointwise output weights method achieved near perfect results only when $\beta$ was chosen well and $N$ was sufficiently large.

\begin{table*}[tpb]
 	\centering
	\begin{tabular}{c  c| ccc | ccc }
	& 	&\multicolumn{3}{c|}{Training} &\multicolumn{3}{c}{Testing}\\
	& N  &Weights (pt) &Weights (bk) &Tensor  &Weights (pt) &Weights (bk) &Tensor\\
	\hline \hline
	\multirow{3}{*}{$\left( \begin{array}{c} f = \sin \\ \beta = 0 \end{array}\right)$}  
&10 &52.01 (0.75) &52.00 (10.88) &100.00 (0.00) &51.88 (0.67) &49.28 (6.77) &100.00 (0.00) \\
%
&20 &53.22 (1.05) &48.90 (13.03) &100.00 (0.00) &53.11 (0.59) &47.68 (7.88) &100.00 (0.00) \\
%
&50 &56.41 (1.56) &60.70 (17.41) &100.00 (0.00) &56.51 (0.95) &61.96 (14.64) &100.00 (0.00) \\
%
\hline
\multirow{3}{*}{$\left( \begin{array}{c} f = \sin \\ \beta = \pi/4 \end{array}\right)$}
%
&10 &88.40 (1.80) &100.00 (0.00) &100.00 (0.00) &87.99 (1.91) &100.00 (0.00) &100.00 (0.00) \\
%
&20 &91.35 (1.72) &100.00 (0.00) &100.00 (0.00) &91.09 (2.05) &100.00 (0.00) &100.00 (0.00) \\
%
&50 &99.51 (0.10) &100.00 (0.00) &100.00 (0.00) &99.49 (0.20) &99.48 (3.68) &100.00 (0.00) \\
%
\hline
\multirow{3}{*}{$\left( \begin{array}{c} f = \tanh \\ \beta = 0 \end{array}\right)$}
&10 &51.77 (0.57) &48.40 (11.49) &100.00 (0.00) &51.86 (0.58) &50.44 (7.67) &100.00 (0.00) \\
&20 &53.09 (0.79) &48.40 (10.62) &100.00 (0.00) &53.27 (0.86) &50.52 (7.45) &100.00 (0.00) \\
&50 &56.25 (1.66) &61.90 (18.98) &100.00 (0.00) &56.39 (1.11) &63.88 (15.10) &100.00 (0.00) \\
\hline
\multirow{3}{*}{$\left(\begin{array}{c} f = \tanh \\ \beta = \pi/4 \end{array} \right)$}
&10 &88.00 (2.13) &100.00 (0.00) &100.00 (0.00) &87.25 (3.62) &100.00 (0.00) &100.00 (0.00) \\
&20 &92.24 (1.50) &100.00 (0.00) &100.00 (0.00) &92.55 (1.53) &100.00 (0.00) &100.00 (0.00) \\
&50 &99.51 (0.11) &100.00 (0.00) &100.00 (0.00) &99.46 (0.14) &100.00 (0.00) &100.00 (0.00) \\
	\end{tabular}
	\caption{Results from `Sine vs.\ Squre Wave' example.  The entries represent the mean and standard deviation (in parenthesis) classification accuracy over 50 trials.  The training and testing classification accuracy are shown for two standard linearout output weight approaches for ESNs, along with the proposed tensor-based classification method.  Several parameters of $N, \beta, f$ are included. }
	\label{tab:SineSquareExample}
\end{table*}

Although~\eqref{eq:tensor class 2} is an instance-based classifier, 100\% classification accuracy on the test set is not guaranteed.  Individual sine or square segments are indeed identical whether from the training or testing set.  However, the resulting reservoir states from these segments are all distinct due to the memory of the reservoir.  That is, the reservoir is not in a resting state when accepting segments in a sequence, and the initial state will continue propagating through the reservoir for some time.  The ouput weight $W_\text{out}$ includes this contamination when it is trained, however the tensor decomposition method can capture the contribution from the initial state in only a small number of factors, while focusing primarily on the reservoir behavior stemming from the input itself.  Overall, the tensor-based approach outperformed the output weights method in that it achieved a higher classification accuracy on the test set for all parameter choices. 

For comparison, this dataset was also studied in~\cite{paquotopto, vandoorne, zhang}, with best reported error rates of $\mathrm{NMSE} \approx 1.5\times 10^{-3}$ in~\cite{paquotopto}, $2.5\%$ misclassifications in~\cite{vandoorne} and $0.3\%$ in~\cite{zhang}.

\subsection{USPS Handwritten Digits}
\begin{figure}[bpht]
	\centering
	\includegraphics{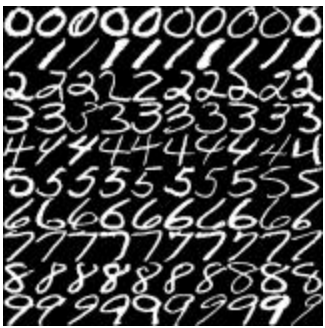}
\caption{Sample digits from the USPS Handwritten Digits dataset.}
\label{fig:USPS example}
\end{figure}

In this collection of experiments, classification is performed on $16\times 16$ grayscale images of handwritten digits.  The dataset is partitioned into 10 classes, representing digits `0' through `9'.   Some samples of these images are shown in Figure~\ref{fig:USPS example}.  
The images are treated as $16\times 16$ spatiotemporal signals, with $y$ coordinates corresponding to the spatial dimension and $x$ coordinates corresponding to the temporal dimension.

For the training dataset, 100 images from each class are randomly selected, and paired with an indicator matrix $y \in \mathbb{R}^{10\times 100}$ where $y(:,j) = e_k$ if the input $A^j$ belongs to the $k^\text{th}$ class.  The test set is formed similarly, but with a distinct 100 images from each class so the test set and training set have no overlap in samples.

The ESN parameters used are ${N\in\{10,25,50,100\}}$, ${f = \tanh}$, ${\beta = \pi/4}$, and the modal ranks for the tensor-based approach are ${J_1 \in \{5, 10, \lfloor N/2\rfloor, \lfloor 3N/4 \rfloor \}}$ and ${J_2 \in \{4, 8, 12\}}$.  For each triplet $(N,J_1,J_2)$, forty simulations are performed with different randomizations in the training and test set selections, as well as the ESN input and reservoir weight matrices for each simulations.  However, for each simulation the same selections are used with each of the classification methods.

For tensor-based classification, $\mathcal{X}\in\mathbb{R}^{N\times 16\times 100}$, the tensor of reservoir states of the training inputs, is decomposed via Algorithm~\ref{alg:HOOI} into the form
\begin{equation*}
	\mathcal{X} \approx \mathcal{F} \times_1 U \times_2 V
\end{equation*}
as in~\eqref{eq:tuckerX}.  
Then a new input $A$ from the test set with reservoir states $X$ is predicted to belong to the $k^\text{th}$ class using Equation~\eqref{eq:tensor class 1}, that is if 
\begin{equation*}
	\left\| X \times_1 U' \times V' - \mathcal{F}(:,:,j)\right\|
\end{equation*}
is minimized for some training input $A^j$ in the $k^\text{th}$ class. 

For trained linear output weights-based classification, a single output weight matrix $W_\text{out} \in \mathbb{R}^{10\times N}$ is generated as in Equation~\eqref{eq:RLS} for each simulation.  Classification is performed on the entire collection of reservoir states for each test input, as in Equation~\eqref{eq:Wout class}.

The results of these simulations are presented in Figure~\ref{alg:HOOI}. Results using the HOOI Algorithm~\ref{alg:HOOI} are displayed in blue, and results using linear output weights are shown in red.   The maximal mean accuracy over all pairs $(J_1,J_2)$ are shown for each $N$ for the tensor-based approach. Note that the tensor-based approach consistently yields higher classification accuracy than the trained linear output weight approach.  Although the results are not competitive with state-of-the-art on this particular dataset, they do show that the tensor-based classification method yields higher results than standard ESN techniques.

\begin{figure*}[htp]
	\centering
	\includegraphics[width=0.98\textwidth]{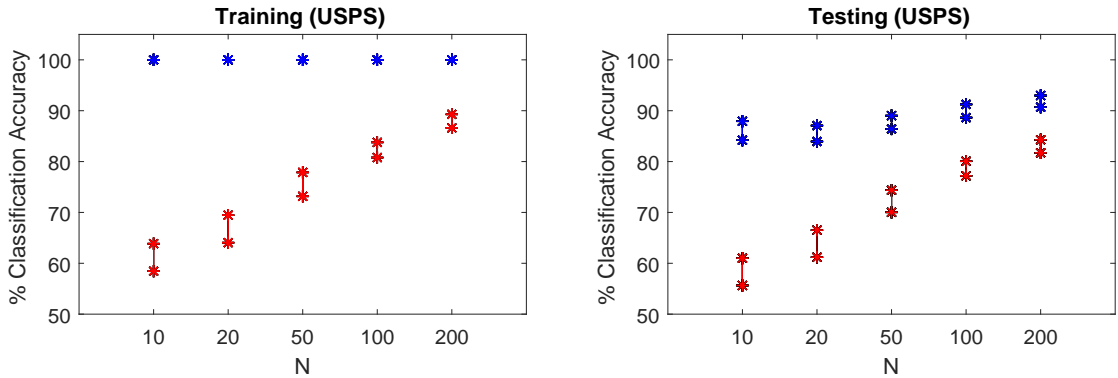}
\caption{Average percent classification accuracy over 40 simulations for each $N$ on the USPS dataset using the tensor-based HOOI Algorithm (in blue) and the linear output weights method (in red).  The vertical lines represent one standard deviation from the mean. }
\label{fig:USPS Results}
\end{figure*}

\subsection{Japanese Vowels}
In this collection of experiments, speaker identification is performed using samples of audio recordings.  The dataset contains 640 samples of nine male speakers saying the Japanese vowel `ae'.  The data is split into 270 training samples of 30 utterances by each speaker, and 370 test samples of 24-88 utterances by each speaker.  Each sample is an $14\times m$ array of cepstrum coefficients, where $m$ is the temporal length of the sample, using 12 cepstrum coefficients and two bias terms.  The dataset, first appearing in~\cite{kudo} and obtained via~\cite{uci}, is popular in machine learning and ESN literature~\cite{barber, guerts, jaegerleaky, prater_NN, strickert}.  A test accuracy of 100\% was reported in~\cite{jaegerleaky} using an ensemble classifier of 1000 ESNs of four leaky-integrator nodes.

In the examples below, we use a single ESN with ${N= 4, 10, 20,}$ or $50$ nodes, a nonlinear activation function ${f = \sin}$ and bias ${\beta = \pi/4}$.   The classifiers are found using~\eqref{eq:RLS} and Algorithm~\ref{alg:HOOI} with~\eqref{eq:tuckerX} from the collection of reservoir states corresponding to the training inputs.  The test inputs are modified by adding Gaussian noise $\sim N(0,\sigma)$ with $\sigma \in \{0.00, 0.05, 0.10\}$.  Classification is then performed on the resulting reservoir states using~\eqref{eq:Wout class} for trained linear output weights or~\eqref{eq:tensor class 1} for the tensor method.  For each pair $(N,\sigma)$, 20 simulations were performed with new randomizations of $W_\text{res}$ and $W_\text{in}$ for each simulation.

\begin{figure*}[tpb]
	\centering
	\includegraphics{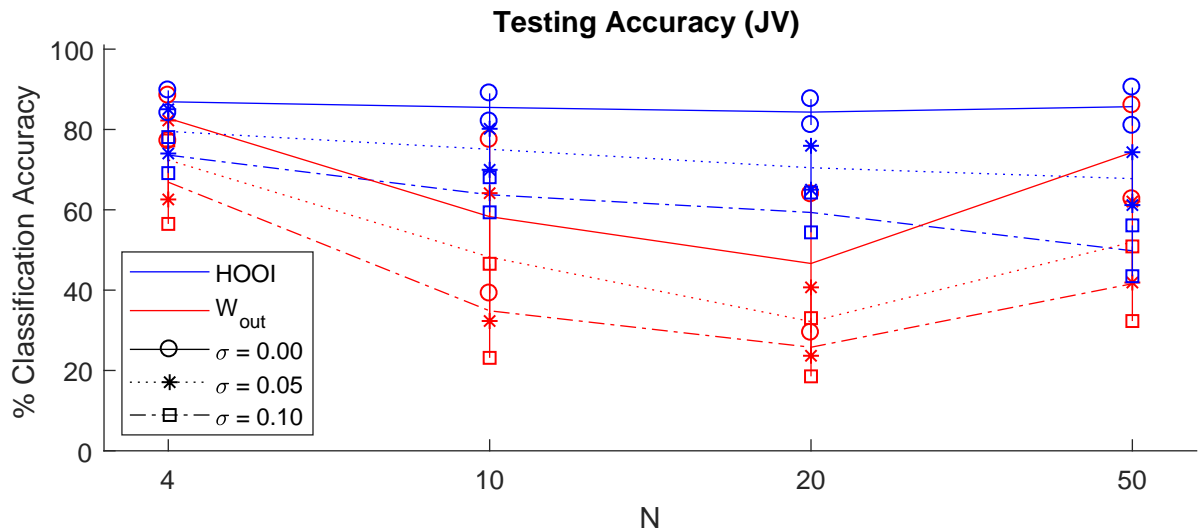}
	\caption{Percent classification test accuracy of the Japanese Vowel dataset.  The results are plotted against $N$, the number of nodes in the reservoir, for both classification methods for several amounts of noise added to the test sets.  The points along the lines represent the mean accuracy over 20 simulations, while the vertical lines represent one standard deviation from the mean.}
	\label{fig:JV}
\end{figure*}

The test classification accuracy results are displayed in Figure~\ref{fig:JV}. In the figure, the blue lines correspond to the tensor-based method, and the red lines correspond to the linear output weight method.  The individual lines within each method correspond to different levels of noise added to the test inputs.  The $x$-axis is the number of reservoir nodes $N$.  The points on the lines give the mean accuracy over the 20 simulations, while the vertical lines represent one standard deviation from the mean.

In the figure, classification accuracy initially decreases as $N$ increases, but eventually improve for large enough $N$.  This is consistent with results published in~\cite{prater_NN}.  Both methods degrade as the level of added noise increases, however the tensor-based method consistently yielded better accuracy results.  Not only does the tensor-based approach have higher mean classification accuracy for all parameter choices, but the standard deviation is smaller indicating that the results are less sensitive to the randomizations in $W_\text{out}$ and $W_\text{in}$.

\section{Conclusion}\label{sec:conclusion}
This work introduced a tensor-based method to perform supervised classification on spatiotemporal data processed in an ESN.  The numerical experiments demonstrate that the proposed method may outperform the traditional trained linear output weights method in terms of classification accuracy.  Future directions include investigating other types of tensor decompositions, including low rank polyadic and sparse decompositions, as well as using other types of non instance-based classifiers on the resulting decompositions.

\subsection*{Acknowledgments}
\noindent This work was cleared for public release by Wright Patterson Air Force Base Public Affairs on 15 Aug 2017.  Case Number: 88ABW-2017-3910.\\


\noindent Any opinions, findings and conclusions or recommendations expressed in this material are those of the author and do not necessarily reflect the view of the United States Air Force.

\bibliographystyle{plain}
\bibliography{mybibfileTensors}

\end{document}